\documentclass[sigconf]{acmart}
\usepackage{booktabs}
\usepackage{graphicx}
\usepackage{tikz}

\setcopyright{none}
\renewcommand\footnotetextcopyrightpermission[1]{}
\acmConference[Preprint]{Preprint}{August 2026}{}
\acmYear{2026}

\AtBeginDocument{%
  }

\title{ReasonCast: Agentic Demand Forecasting with Selective Semantic Reasoning}

\author{Ziyue Yang}
\authornote{Ziyue Yang and Chaolin Xu contributed equally to this work.}
\affiliation{%
  \institution{Taobao and Tmall Group, Alibaba Group}
  \country{China}}

\author{Chaolin Xu}
\authornotemark[1]
\affiliation{%
  \institution{Taobao and Tmall Group, Alibaba Group}
  \country{China}}

\author{Yijing Wang}
\affiliation{%
  \institution{Taobao and Tmall Group, Alibaba Group}
  \country{China}}

\author{Tiankai Gu}
\affiliation{%
  \institution{Taobao and Tmall Group, Alibaba Group}
  \country{China}}

\author{Hui Yang}
\affiliation{%
  \institution{Taobao and Tmall Group, Alibaba Group}
  \country{China}}

\author{Yanhong Lin}
\affiliation{%
  \institution{Taobao and Tmall Group, Alibaba Group}
  \country{China}}

\author{Kaiyuan Liu}
\affiliation{%
  \institution{Taobao and Tmall Group, Alibaba Group}
  \country{China}}

\author{Fei Xiao}
\authornote{Corresponding author.}
\affiliation{%
  \institution{Taobao and Tmall Group, Alibaba Group}
  \country{China}}

\begin{document}

\begin{abstract}
Demand forecasting increasingly requires combining two complementary sources of information: historical sales reveal recurring numerical dynamics, while future promotions, holidays, price changes, and platform interventions provide forward-looking knowledge that may not be recoverable from past observations alone. Time-series foundation models are strong numerical forecasters, and large language models can interpret heterogeneous event context. However, existing text-enhanced forecasting methods often encode such context into generic representations and fuse it uniformly with time-series features, without explicitly distinguishing which semantic effects are forecast-relevant, where they should interact with temporal representations, or how they should modify future dynamics.

We introduce ReasonCast, a structured semantic intervention framework that translates event knowledge into forecast-specific operations. Before constructing an intervention, a lightweight agent examines the event context, the no-text forecast, and its predictive uncertainty to determine whether textual reasoning is needed and whether additional evidence or temporal-statistics tools should be invoked. For known regime-changing events, including holidays and mega-sales, semantic reasoning is activated by default. Rather than injecting free-form text, ReasonCast represents event knowledge through structured fields describing event relevance, demand direction, temporal shape, amplitude, and peak intensity. These fields interact selectively with event-related temporal components of a pretrained time-series foundation model. An additive path corrects local trends and temporal shapes, a multiplicative path captures event-driven level shifts, and an instance-wise mechanism calibrates intervention strength. When semantic intervention is unnecessary, the original numerical forecast can be preserved exactly.

To make the resulting semantics not only well-formed but forecast-effective, ReasonCast introduces a forecast-grounded post-training curriculum. Schema SFT establishes valid and internally consistent semantic fields; semantic-field RL calibrates direction, shape, amplitude, and peak judgments using forecast-derived supervision; and forecast-utility RL evaluates candidate semantic interventions through a frozen forecaster, aligning reasoning outputs with marginal forecast improvement while penalizing negative transfer and unnecessary tool cost. ReasonCast lowers WMAPE by 3.29, 1.25, and 0.47 percentage points on holiday-sensitive categories, mega-sale-sensitive categories, and M5 event windows, respectively. On stable-sales periods, indiscriminate semantic intervention increases WMAPE by 1.68 percentage points, whereas suppressing unnecessary intervention preserves the numerical backbone. These results demonstrate the value of moving beyond generic text fusion toward structured, forecast-aligned semantic intervention.
\end{abstract}

\keywords{time series forecasting, demand forecasting, large language models, agentic reasoning, multimodal learning, text--time-series fusion, event-aware forecasting}

\maketitle

\section{Introduction}
Commerce demand reflects two fundamentally different sources of variation. Recurring regularities---such as trend, seasonality, autocorrelation, and repeated campaign patterns---can be learned from historical observations. Future promotions, holidays, price changes, platform interventions, and product-specific events, however, may cause shifts that are not identifiable from past sales alone, a distinction also reflected in demand benchmarks and models of known future covariates~\cite{makridakis2022m5,lim2021tft,wang2024timexer}. Time-series foundation models (TSFMs) provide strong numerical backbones by transferring recurring structure across large collections of series~\cite{das2024timesfm,ansari2024chronos,woo2024moirai,liu2024timer,goswami2024moment}, with recent work further scaling model capacity, context length, and pattern specialization~\cite{liu2025moiraimoe,liu2025sundial,liu2025timerxl}. Yet a numerical backbone cannot infer a future event that is absent from its inputs.

Large language models (LLMs) offer a complementary capability: they can interpret heterogeneous event records and convert contextual knowledge into forecasting-relevant judgments. The value of such context is nevertheless conditional. Text helps when it supplies future events, constraints, or background information unavailable from numerical history~\cite{williams2025context}; broad empirical evidence further shows that multimodal gains depend strongly on whether text contributes signals not already captured by the temporal input and backbone~\cite{zhang2026multimodality}. Event-aware studies likewise find that unfiltered news can degrade forecasting relative to carefully selected evidence~\cite{wang2024newsforecast}. Thus, the central question is not simply how to fuse more text, but \emph{whether, where, and to what extent semantic reasoning should alter a strong numerical forecast}.

Representative forecasting designs, summarized in Table~\ref{tab:paradigm-comparison}, leave three challenges unresolved. First, semantic value is heterogeneous across products, events, and forecast windows, requiring \emph{instance-adaptive control}. Second, useful semantics must enter the computation at an appropriate location and in an appropriate form: some events change local trends and shapes, whereas major campaigns produce multiplicative level shifts. Third, language supervision rewards plausible reasoning but does not ensure downstream predictive value. These challenges call for a forecaster that separates semantic reasoning from numerical prediction and treats text as a controlled intervention rather than an always-on modality.

\begin{table*}[!t]
\centering
\caption{Comparison of representative forecasting designs. $\checkmark$, $\triangle$, and $\times$ denote explicit, partial or method-dependent, and no explicit support. Reasoning over exogenous events goes beyond encoding external text to infer forecast-relevant effects. Selective semantic intervention combines instance-level \textsc{Skip}/\textsc{Basic}/\textsc{Tool} routing with continuous control over intervention strength; designs without an explicit abstention route provide only partial support. Reasoner adaptation covers both parametric post-training and non-parametric use of forecasting feedback.}
\label{tab:paradigm-comparison}
\scriptsize
\setlength{\tabcolsep}{3.0pt}
\renewcommand{\arraystretch}{1.18}
\begin{tabular}{p{0.24\textwidth}ccccp{0.19\textwidth}}
\toprule
\textbf{Forecasting design} & \shortstack{\textbf{Numerical}\\\textbf{forecast owner}} & \shortstack{\textbf{Reasoning over}\\\textbf{exogenous events}} & \shortstack{\textbf{Selective semantic}\\\textbf{intervention}} & \shortstack{\textbf{Exact no-text}\\\textbf{preservation}} & \textbf{Reasoner adaptation} \\
\midrule
TS-only / TSFM~\cite{ansari2024chronos,das2024timesfm} & TSFM & $\times$ & $\times$ & -- & N/A \\
LLM numerical forecasting~\cite{gruver2023llmtime,jin2024timellm} & LLM & $\times$ & $\times$ & $\times$ & Frozen / prompting \\
Joint multimodal forecasting~\cite{jia2024gpt4mts,li2026tats,wang2025chattime,wu2026aurora,zhou2025balmtsf} & Joint model & $\triangle$ & $\times$ & $\times$ & Task-level training / SFT \\
From News to Forecast~\cite{wang2024newsforecast} & LLM & $\checkmark$ & $\times$ & $\times$ & Reflection + evidence reselection \\
VoT~\cite{wang2026vot} & Hybrid branches & $\checkmark$ & $\triangle$ & $\times$ & Retrieval + historical ICL \\
\textbf{ReasonCast} & \textbf{TSFM} & $\checkmark$ & $\checkmark$ & $\checkmark$ & \shortstack{\textbf{SFT + semantic-field RL}\\\textbf{+ forecast-utility RL}} \\
\bottomrule
\end{tabular}
\end{table*}

We introduce \textbf{ReasonCast}, an agentic demand forecasting framework following a \emph{route--reason--intervene--forecast} paradigm. Given event context, the no-text forecast, and its entropy, the agent selects $\textsc{Skip}$, low-cost $\textsc{Basic}$ semantics, or tool-augmented $\textsc{Tool}$ semantics; holidays and mega-sales rule out $\textsc{Skip}$. A pretrained TSFM remains responsible for numerical prediction.

For non-skip actions, event-related temporal components alone query semantic tokens. An instance-wise gate controls intervention strength, while additive and multiplicative paths model local and level changes. $\textsc{Skip}$ disables both paths and exactly recovers the no-text computation. Alignment pretraining grounds event semantics, and forecast-aligned policy optimization rewards routing and outputs by their marginal improvement over the backbone.

We evaluate ReasonCast on large-scale commerce data at multiple granularities and public item-level demand benchmarks, separating normal and event-driven periods and stress-testing corrupted context. Our contributions are:
\begin{itemize}
    \item We formulate event-enhanced demand forecasting as \emph{selective semantic intervention}, shifting the focus from extracting more text to controlling its marginal influence on a strong numerical forecaster.
    \item We propose a three-action agent policy that routes each instance to no intervention, low-cost structured reasoning, or tool-augmented reasoning, with mandatory semantic activation for holidays and mega-sales.
    \item We combine selective event interaction, per-instance gating, additive and multiplicative corrections, and exact recovery of the no-text backbone whenever the agent skips intervention.
\end{itemize}

\section{Related Work}
\subsection{Numerical and Language-Based Forecasting}
Numerical forecasting has progressed from probabilistic and covariate-aware
models~\cite{salinas2020deepar,lim2021tft} to specialized linear, patch-based,
inverted, and multiscale architectures~\cite{zeng2023dlinear,nie2023patchtst,liu2024itransformer,wang2024timemixer,wang2024timexer}. Time-series foundation models further transfer recurring structure across datasets through value tokenization, decoder-style generation, or heterogeneous-series pretraining~\cite{das2024timesfm,ansari2024chronos,woo2024moirai,liu2024timer,goswami2024moment,liu2025moiraimoe}. These models are strong numerical specialists, but they cannot infer a future event absent from their inputs.

Other work repurposes pretrained language models as numerical forecasters or representation learners~\cite{gruver2023llmtime,zhou2023onefitsall,sun2024test,jin2024timellm,cao2024tempo,liu2024autotimes,liu2024unitime}. Because language components do not consistently explain the resulting gains~\cite{tan2024language}, ReasonCast retains a TSFM as forecast owner and assigns the LLM a distinct role: reasoning over event knowledge rather than generating sales values.

\subsection{Multimodal Forecasting and Cross-Modal Alignment}
Text-enhanced forecasting uses either endogenous descriptions derived from the series~\cite{jin2024timellm,sun2024test,liu2025calf,zhou2025balmtsf,jia2026m3time} or exogenous evidence such as news, events, and domain conditions~\cite{jia2024gpt4mts,liu2024timemmd,li2026tats,wu2026aurora,wang2025chattime}. The former can improve representation transfer but may duplicate numerical information; the latter can reveal shifts unavailable from history. Empirical studies accordingly find that multimodal gains depend on backbone capacity, alignment, data scale, and whether text adds predictive information~\cite{williams2025context,zhang2026multimodality}. ReasonCast turns this conditional utility into an instance-level control objective.

\subsection{Event Reasoning and Forecast-Aware Agents}
Language agents can interleave reasoning, actions, and tool use~\cite{yao2023react,schick2023toolformer}. In forecasting, From News to Forecast refines evidence selection using validation feedback~\cite{wang2024newsforecast}, while VoT combines event and numerical branches with retrieval and adaptive fusion~\cite{wang2026vot}. ReasonCast instead routes each instance among no intervention, low-cost reasoning, and tool-augmented reasoning; structured semantics modify a dedicated TSFM only when selected. Forecast-aligned optimization trains this policy for marginal utility, and the no-text forecast is retained exactly when the agent skips.

\section{Problem Formulation}
For each item or leaf category $i$, let
\begin{equation}
  \mathbf{x}_i = [x_{i,1},\ldots,x_{i,L}]
\end{equation}
denote an observed demand history of length $L$, and let $\mathbf{c}_i$ denote the event records and contextual information available before the forecast origin. The goal is to predict the next $H=7$ observations,
\begin{equation}
  \hat{\mathbf{y}}_i = [\hat{y}_{i,1},\ldots,\hat{y}_{i,H}].
\end{equation}
A pretrained Chronos-2 model $f_{\theta}$~\cite{ansari2025chronos2} provides the no-text forecast $\hat{\mathbf{y}}^{\mathrm{ts}}_i=f_{\theta}(\mathbf{x}_i)$ and entropy $\mathcal{H}_i$. With $\mathbf{o}_i=(\mathbf{c}_i,\hat{\mathbf{y}}^{\mathrm{ts}}_i,\mathcal{H}_i)$, the agent chooses
\begin{equation}
  a_i\sim\pi_{\phi}^{\mathrm{route}}(\mathbf{o}_i),\quad
  a_i\in\{\textsc{Skip},\textsc{Basic},\textsc{Tool}\},
  \label{eq:agent-route}
\end{equation}
where holidays and mega-sales enforce $a_i\neq\textsc{Skip}$. We set $\mathbf{T}_i=\varnothing$ for $\textsc{Skip}$ and otherwise generate semantic tokens with $\pi_{\phi}^{\mathrm{sem}}(\mathbf{o}_i,a_i)$. The forecast $F_{\theta,\psi}(\mathbf{x}_i,\mathbf{T}_i)$ satisfies
\begin{equation}
  a_i=\textsc{Skip}\Longrightarrow
  F_{\theta,\psi}(\mathbf{x}_i,\varnothing)=f_{\theta}(\mathbf{x}_i).
\end{equation}

\section{ReasonCast}
\subsection{Framework Overview}
\begin{figure*}[t]
  \centering
  \includegraphics[width=0.98\textwidth]{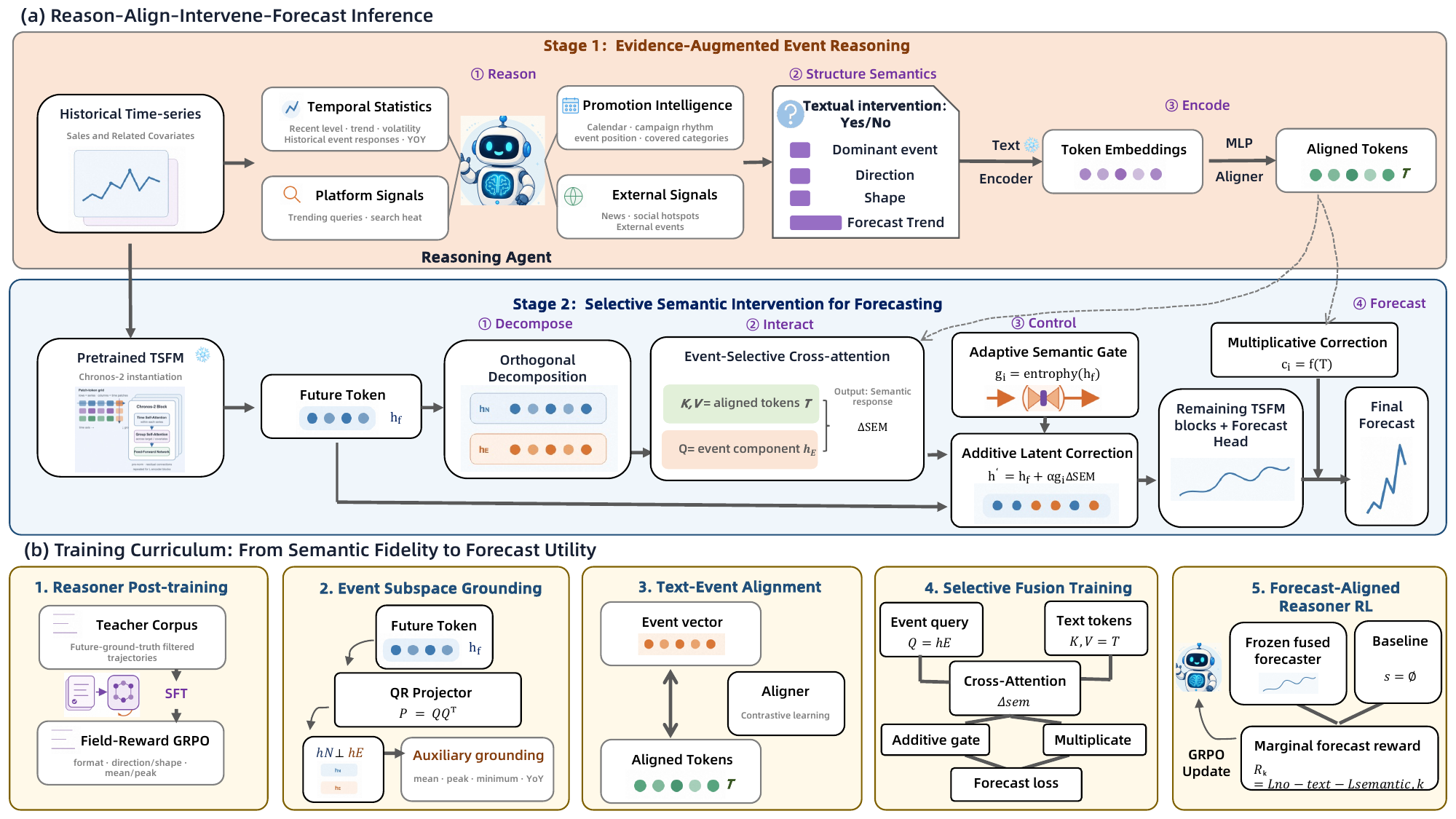}
  \caption{Overview of ReasonCast. \textbf{(a)} ReasonCast follows a reason--align--intervene--forecast pipeline. An evidence-augmented LLM agent combines the historical series with temporal statistics, promotion intelligence, platform signals, and external signals to produce structured event semantics. A text encoder and an MLP aligner map these semantics to aligned tokens $\mathbf{T}$. In parallel, a pretrained time-series foundation model (TSFM) produces a future token $\mathbf{h}_F$, which is orthogonally decomposed at the fusion interface into event-related and complementary components $\mathbf{h}_E$ and $\mathbf{h}_N$. Only $\mathbf{h}_E$ queries $\mathbf{T}$ through event-selective cross-attention. An adaptive semantic gate modulates the additive latent correction for local trend and shape changes, while a complementary multiplicative path handles large event-driven level shifts. Without semantic context, both correction paths are disabled and the fusion interface exactly recovers the original TSFM computation. \textbf{(b)} The training curriculum comprises schema SFT, semantic-field RL, orthogonal event-subspace grounding and selective fusion training, and forecast-utility RL. During forecast-utility RL, candidate semantics are compared with the cached semantic-field-policy fusion result, while degradation beyond the no-text TSFM is explicitly penalized.}
  \Description{A two-panel overview. The first panel traces event evidence
  through an LLM reasoner, text alignment, event-selective fusion, and a
  time-series forecast. The second panel shows the SFT, semantic-field RL,
  fusion training, and forecast-utility RL curriculum.}
  \label{fig:overview}
\end{figure*}

ReasonCast is trained with a three-phase curriculum. First, schema SFT establishes valid structured outputs, after which semantic-field RL calibrates forecast-derived direction, temporal shape, amplitude, and peak fields. Second, the semantic-field policy regenerates the fusion-training corpus, on which we learn orthogonal event-subspace grounding, multimodal alignment, and selective dual-path fusion. Third, the trained forecaster is frozen and reused as a stable downstream evaluator for forecast-utility RL, which jointly optimizes routing and semantic outputs without updating the numerical model. Figure~\ref{fig:overview} summarizes the framework.

\subsection{Agentic Event Routing and Semantic Reasoning}
$\textsc{Skip}$ terminates semantic processing, $\textsc{Basic}$ uses available context, and $\textsc{Tool}$ retrieves evidence or invokes temporal-statistics tools for ambiguous or high-entropy cases. Hard-triggered holidays and mega-sales use at least $\textsc{Basic}$. For either non-skip action, Qwen3-32B~\cite{yang2025qwen3} produces
\begin{equation}
  \mathbf{s}_i =
  (r_i,\,e_i,\,d_i,\,q_i,\,u_i,\,p_i,\,z_i),
\end{equation}
where $r_i$ denotes event relevance, $e_i$ the dominant event, $d_i$ demand direction, $q_i$ temporal shape, $u_i$ the amplitude multiplier, $p_i$ the peak multiplier, and $z_i$ a concise forecast-trend narrative. A text encoder and MLP aligner produce $D$-dimensional tokens $\mathbf{T}_i$. SFT teaches the schema; forecast-aligned optimization calibrates routing and semantic utility. The agent decides whether and how much to reason, while the TSFM retains numerical forecasting.

\paragraph{Reasoner supervision.} We stratify 40,000 item--window candidates by trajectory regime and obtain 11,627 teacher generations. Deterministic field-consistency filters retain 7,312 responses, which are compressed, deduplicated, and rebalanced into 4,590 SFT examples; Appendix~\ref{app:reasoner-details} provides the full construction protocol.

\subsection{Event-Selective Temporal Interaction}
Let $\mathbf{H}_i\in\mathbb{R}^{M\times D}$ be temporal tokens at a fusion layer. From a learned basis $\mathbf{B}\in\mathbb{R}^{D\times d_E}$, QR factorization yields an orthonormal event basis $\mathbf{Q}$ with $\mathbf{Q}^{\top}\mathbf{Q}=\mathbf{I}$. We define the orthogonal projector $\mathbf{P}_E=\mathbf{Q}\mathbf{Q}^{\top}$ and decompose
\begin{equation}
  \mathbf{H}^{E}_i=\mathbf{H}_i\mathbf{P}_E, \qquad
  \mathbf{H}^{N}_i=\mathbf{H}_i(\mathbf{I}-\mathbf{P}_E).
\end{equation}
Because $\mathbf{P}_E^{\top}=\mathbf{P}_E$ and $\mathbf{P}_E^2=\mathbf{P}_E$, the decomposition guarantees $\mathbf{H}^{E}_i+\mathbf{H}^{N}_i=\mathbf{H}_i$ and $\langle\mathbf{H}^{E}_i,\mathbf{H}^{N}_i\rangle_F=0$. Textual interaction is computed with the event component as query:
\begin{equation}
  \Delta\mathbf{H}_i =
  \operatorname{CrossAttn}
  (Q=\mathbf{H}^{E}_i,K=\mathbf{T}_i,V=\mathbf{T}_i).
\end{equation}
The fused representation is
\begin{equation}
  \widetilde{\mathbf{H}}_i
  =\mathbf{H}^{N}_i+\mathbf{H}^{E}_i
  +m_i\alpha g_i\Delta\mathbf{H}_i,
  \label{eq:additive}
\end{equation}
where $m_i$ is a text-availability mask, $\alpha$ is a residual scale, and $g_i$ is an instance-wise gate. The interface is applied to the Chronos-2 future-token representation before the prediction head. Thus, only the event component queries text; because the attention output is not reprojected, we make no stronger claim that the semantic update itself remains in the event component.

\subsection{Multimodal Alignment Pretraining}
Before end-to-end forecasting training, we align pooled event semantics $\mathbf{z}^{T}_i$ with event-related temporal representations $\mathbf{z}^{E}_i$. Motivated by contrastive cross-modal alignment~\cite{jia2021align} and its recent use in language--time-series forecasting~\cite{liu2025calf,zhou2025balmtsf}, a contrastive objective encourages matched event--series pairs to be close and mismatched pairs to remain separated:
\begin{equation}
  \mathcal{L}_{\mathrm{align}}
  =-\frac{1}{B}\sum_{i=1}^{B}
  \log
  \frac{\exp(\operatorname{sim}(\mathbf{z}^{E}_i,\mathbf{z}^{T}_i)/\tau)}
  {\sum_{j=1}^{B}\exp(\operatorname{sim}(\mathbf{z}^{E}_i,\mathbf{z}^{T}_j)/\tau)}.
\end{equation}
The denominator uses the other text embeddings in the same minibatch as negatives.

\subsection{Instance-Adaptive Dual-Path Correction}
The value of semantic context varies across items and forecast windows. ReasonCast uses an instance-wise gate
\begin{equation}
  g_i=g_{\max}\sigma\!\left(\operatorname{MLP}(\mathbf{u}_i)\right),
\end{equation}
where $\mathbf{u}_i$ concatenates pooled semantic features, recent-demand volatility, entropy of the no-text forecast, and encoded event descriptors, all available at inference time. The gate modulates the additive latent residual in Eq.~\eqref{eq:additive}.

Latent residuals are well suited to fine-grained temporal pattern changes but may have limited authority over large level shifts. We therefore introduce a complementary multiplicative correction:
\begin{equation}
  \hat{\mathbf{y}}_i
  =\hat{\mathbf{y}}^{\mathrm{lat}}_i
  \odot
  \exp\!\left(m_i\beta g_i\boldsymbol{\rho}_i\right),
  \label{eq:multiplicative}
\end{equation}
where $\hat{\mathbf{y}}^{\mathrm{lat}}_i$ is decoded from the additively fused representation, $\boldsymbol{\rho}_i$ is a bounded semantic level-shift signal, and $\beta$ controls the correction scale.

\subsection{Exact No-Text Identity}
When the agent selects $\textsc{Skip}$, or when textual context is unavailable, $m_i=0$. Equations~\eqref{eq:additive} and~\eqref{eq:multiplicative} then reduce to
\begin{equation}
  \widetilde{\mathbf{H}}_i=\mathbf{H}_i,
  \qquad
  \hat{\mathbf{y}}_i=\hat{\mathbf{y}}^{\mathrm{ts}}_i.
\end{equation}
The property is enforced by the routing action and an explicit mask rather than learned approximately. Consequently, a skipped stable-sales instance recovers the Chronos-2 backbone prediction exactly. The guarantee applies to the fusion interface. Although its event subspace is geometrically orthogonal to the complement, this constraint does not make the subspace causally or globally identifiable.

\subsection{Forecast-Aligned Reasoner Post-Training}
\label{sec:two-stage-grpo}
Language supervision alone can yield plausible but forecast-irrelevant text,
whereas direct loss optimization may reward malformed semantics that happen to
improve a prediction. Following schema SFT, we optimize the reasoner with two RL stages: semantic-field RL calibrates verifiable semantic fields, and forecast-utility RL optimizes their downstream value against a frozen forecaster.

\paragraph{Semantic-field RL: structured semantic calibration.}
For prompt $i$, GRPO samples $G$ structured responses. Forecast-derived targets
score direction, temporal shape, amplitude, and peak:
\begin{equation}
  A_{ij}=\sum_{k\in\{d,q,a,p\}}w_k a^k_{ij}
  -C_{ij}^{\mathrm{struct}},
  \label{eq:stage1-field-reward}
\end{equation}
where $C_{ij}^{\mathrm{struct}}$ penalizes invalid formats and cross-field
contradictions. Group-normalized rewards and a KL reference to the SFT policy
yield the field-calibrated reasoner $\pi_1$.

\paragraph{Forecast-utility RL: frozen-forecaster optimization.}
We use $\pi_1$ to regenerate the fusion corpus, train the alignment and
dual-path forecaster $F_1$, and then freeze it. Each rollout selects
$\textsc{Skip}$, $\textsc{Basic}$, or $\textsc{Tool}$; holidays and mega-sales
disallow $\textsc{Skip}$. Non-skip candidates must remain parseable,
cross-field consistent, and close to the cached semantic-field score before
being evaluated by $F_1$.

Let $M_{ij}$ be candidate MAE, $M_i^{\mathrm{ref}}$ the cached semantic-field-policy fusion
MAE, $M_i^{\mathrm{ts}}$ the no-text MAE, and $S_i$ a sample-specific scale. We
measure downstream improvement and one-sided harm as
\begin{equation}
  D_{ij}=\frac{M_i^{\mathrm{ref}}-M_{ij}}{S_i},\qquad
  H_{ij}=\frac{[M_{ij}-(1+\eta)M_i^{\mathrm{ts}}]_+}{S_i}.
  \label{eq:stage2-utility}
\end{equation}
With semantic-validity indicator $v_{ij}$, semantic margin $C_{ij}$, and route
cost $c(a_{ij})$, the reward is
\begin{equation}
  R_{ij}=\begin{cases}
    b+\lambda_u\operatorname{clip}(D_{ij}), & a_{ij}=\textsc{Skip},\\
    A_{ij}-\gamma, & v_{ij}=0,\\
    b+\lambda_c C_{ij}+\lambda_u\operatorname{clip}(D_{ij}-\lambda_hH_{ij})
      -\lambda_{\mathrm{cost}}c(a_{ij}), & v_{ij}=1.
  \end{cases}
  \label{eq:stage2-total-reward}
\end{equation}
Thus, abstention competes directly with semantic intervention, while tool use
must justify its additional cost. GRPO is initialized from $\pi_1$ and uses
the semantic-field policy globally, plus SFT conditionally, as KL references to preserve semantic
fidelity. Appendix~\ref{app:reasoner-details} provides the validity threshold,
reference conditions, and optimization settings.

\section{Experiments}
\subsection{Research Questions}
We evaluate ReasonCast through four questions:
\begin{itemize}
  \item \textbf{RQ1:} Does ReasonCast improve event-period forecasting and
  transfer to M5 while preserving stable-period performance?
  \item \textbf{RQ2:} Do discrete routing and entropy-conditioned gating
  allocate semantic intervention more effectively than always-on, ungated, or
  globally weighted fusion?
  \item \textbf{RQ3:} What do orthogonal decomposition, alignment, gating, and
  additive--multiplicative correction each contribute?
  \item \textbf{RQ4:} How do SFT, semantic-field RL, and forecast-utility
  RL affect semantic fidelity and downstream forecast utility?
\end{itemize}

\subsection{Datasets and Evaluation Protocol}
The proprietary commerce dataset contains 15,696 daily sales series at the leaf-category level, spanning August 10, 2024 to May 30, 2026. We adopt a chronological split: August 10, 2024 to August 31, 2025 for training, September 1 to October 31, 2025 for model selection, and November 1, 2025 to May 30, 2026 for evaluation. The test set comprises 15 seven-day forecasting windows, including three mega-sale, ten holiday, and two stable-sales windows. To prevent temporal leakage, all textual descriptions and event information are restricted to records available before the corresponding forecast origin. For M5~\cite{makridakis2022m5}, we evaluate 3,049 item-level series aggregated across all ten stores over 14 seven-day event-centered windows, using only calendar, event, and price covariates available at each forecast origin.

\subsection{Baselines}
We compare Seasonal Naive with LightGBM~\cite{ke2017lightgbm}, PatchTST~\cite{nie2023patchtst}, and iTransformer~\cite{liu2024itransformer}; pretrained TimesFM~\cite{das2024timesfm}, Chronos~\cite{ansari2024chronos}, and Chronos-2 small~\cite{ansari2025chronos2}; and the text-enhanced Time-LLM~\cite{jin2024timellm} and VoT~\cite{wang2026vot}. Chronos-2 small is evaluated zero-shot and with the Uni-FT and Multi-FT variants shown in the tables. Numerical models receive historical demand, while covariate-enabled and text-enhanced variants additionally receive only calendar, event, price, or text fields available before the forecast origin. All methods use the same split, rolling origins, seven-day horizon, and metrics; trainable baselines are selected by validation WMAPE without test-window tuning.

\subsection{Metrics}
We evaluate point forecasting accuracy using weighted mean absolute percentage
error (WMAPE) and mean absolute error (MAE). Let $\mathcal{D}_s$ denote the
set of leaf-category--day observations in an evaluation slice $s$. We compute
\begin{equation}
  \operatorname{WMAPE}_s
  =
  \frac{\sum_{(i,t)\in\mathcal{D}_s}
  |y_{i,t}-\hat{y}_{i,t}|}
  {\sum_{(i,t)\in\mathcal{D}_s}|y_{i,t}|},
\end{equation}
and
\begin{equation}
  \operatorname{MAE}_s
  =
  \frac{1}{|\mathcal{D}_s|}
  \sum_{(i,t)\in\mathcal{D}_s}
  |y_{i,t}-\hat{y}_{i,t}|.
\end{equation}
WMAPE is the primary metric because it measures aggregate forecasting error
relative to realized demand and places greater emphasis on high-volume
observations. MAE complements WMAPE by retaining the absolute magnitude of
forecasting errors. Both metrics are computed separately for each evaluation
slice, including stable-sales periods, holiday and mega-sale periods,
event-sensitive and event-insensitive leaf categories, and the M5 dataset.
Lower values indicate better performance.

\subsection{Main Results}
Table~\ref{tab:main-event} compares event-centered performance on the
proprietary dataset and M5. Holiday and mega-sale categories are split into
sensitive and insensitive groups using training data only; M5 uses windows
around major calendar events. Table~\ref{tab:stable-sales} separately tests the
boundary case in which regular temporal dynamics already explain most demand.

\begin{table*}[!t]
  \centering
  \caption{Main forecasting results on event-centered evaluations. Cells report
  WMAPE/MAE; lower values are better. Sens. and Insens. denote event-sensitive
  and event-insensitive leaf categories, respectively. The best results are
  highlighted in bold, and the second-best results are underlined.}
  \label{tab:main-event}
  \setlength{\tabcolsep}{2.2pt}
  \scriptsize
  \resizebox{\textwidth}{!}{%
  \begin{tabular}{l*{14}{c}}
    \toprule
    Method
      & \multicolumn{6}{c}{Holiday}
      & \multicolumn{6}{c}{Mega-sale}
      & \multicolumn{2}{c}{M5 event windows} \\
    \cmidrule(lr){2-7}
    \cmidrule(lr){8-13}
    \cmidrule(lr){14-15}
      & \multicolumn{2}{c}{Sens.}
      & \multicolumn{2}{c}{Insens.}
      & \multicolumn{2}{c}{All}
      & \multicolumn{2}{c}{Sens.}
      & \multicolumn{2}{c}{Insens.}
      & \multicolumn{2}{c}{All}
      & WMAPE & MAE \\
    \cmidrule(lr){2-3}
    \cmidrule(lr){4-5}
    \cmidrule(lr){6-7}
    \cmidrule(lr){8-9}
    \cmidrule(lr){10-11}
    \cmidrule(lr){12-13}
    \cmidrule(lr){14-15}
    \midrule
    Seasonal Naive
      & 119.96\% & 2738.47 & 56.69\% & 609.09 & 47.39\% & 571.90
      & 27.94\% & 766.85 & 12.49\% & 283.57 & 16.55\% & 434.69
      & 44.55\% & 5.47 \\
    LightGBM
      & 74.61\% & 1671.02 & 43.20\% & 489.11 & 34.23\% & 447.08
      & 26.94\% & 732.37 & 12.83\% & 290.49 & 16.23\% & 425.58
      & 34.06\% & 4.17 \\
    PatchTST
      & 84.90\% & 1771.26 & 47.80\% & 516.75 & 39.78\% & 485.47
      & 22.94\% & 652.38 & \underline{9.91\%} & \textbf{214.34}
      & 13.41\% & 355.74 & 48.31\% & 5.91 \\
    iTransformer
      & 79.29\% & 1849.54 & 42.03\% & 486.83 & 33.64\% & 450.65
      & 23.07\% & 653.62 & 11.30\% & 248.18 & 14.49\% & 382.73
      & 34.39\% & 4.21 \\
    \midrule
    TimesFM
      & 67.23\% & 1586.46 & 48.82\% & 557.09 & 39.37\% & 517.45
      & 24.20\% & 701.77 & 14.57\% & 316.56 & 17.63\% & 465.66
      & 34.63\% & 4.23 \\
    Chronos1
      & 44.54\% & \underline{1163.21} & 37.25\% & 450.04 & 29.70\% & 417.03
      & 26.99\% & 730.93 & 10.97\% & 248.37 & 14.90\% & 391.61
      & 35.38\% & 4.33 \\
    Chronos2-small
      & 55.42\% & 1412.40 & 38.50\% & 464.16 & 31.28\% & 437.37
      & 24.59\% & 681.33 & 10.83\% & 245.81 & 14.64\% & 389.11
      & 34.40\% & 4.20 \\
    Chronos2-small (Uni-FT)
      & \underline{42.39\%} & 1185.03 & 34.94\% & 415.31 & 27.89\% & 388.79
      & 25.20\% & 694.97 & 10.75\% & 242.43 & 14.55\% & 382.88
      & 33.21\% & 4.08 \\
    \midrule
    Chronos2-small (Multi-FT)
      & 45.30\% & 1268.46 & \underline{31.64\%} & \underline{390.67}
      & \underline{25.59\%} & \underline{370.86}
      & \underline{22.14\%} & \underline{622.10}
      & 10.02\% & 222.37 & \underline{13.35\%} & \underline{352.24}
      & \underline{33.10\%} & \underline{4.06} \\
    Time-LLM
      & 87.76\% & 1972.79 & 57.31\% & 602.99 & 47.03\% & 556.71
      & 25.89\% & 721.11 & 13.57\% & 306.44 & 16.97\% & 445.65
      & 38.99\% & 4.78 \\
    VoT
      & 61.51\% & 1514.34 & 32.38\% & 417.44 & 27.33\% & 400.96
      & 24.74\% & 711.91 & 13.42\% & 295.53 & 16.83\% & 444.76
      & 38.05\% & 4.69 \\
    ReasonCast (agent-routed)
      & \textbf{39.10\%} & \textbf{1080.71}
      & \textbf{25.08\%} & \textbf{331.09}
      & \textbf{21.06\%} & \textbf{322.24}
      & \textbf{20.89\%} & \textbf{589.94}
      & \textbf{9.71\%} & \underline{215.61}
      & \textbf{12.87\%} & \textbf{339.68}
      & \textbf{32.63\%} & \textbf{4.00} \\
    \bottomrule
  \end{tabular}%
  }
\end{table*}

ReasonCast obtains the lowest WMAPE on all six proprietary event slices and on
M5 event windows, with the largest margins during holidays. It also achieves
the best WMAPE and second-best MAE on mega-sale-insensitive categories,
indicating that the gain transfers beyond the most responsive groups and beyond
the proprietary dataset.

\begin{table}[!t]
  \centering
  \caption{Forecasting performance during stable-sales periods on the
  proprietary commerce dataset. Lower values are better. The best results are
  highlighted in bold, and the second-best results are underlined.}
  \label{tab:stable-sales}
  \begin{tabular}{lcc}
    \toprule
    Method & WMAPE & MAE \\
    \midrule
    Seasonal Naive & 10.89\% & 254.82 \\
    LightGBM & 11.10\% & 259.88 \\
    PatchTST & 9.72\% & 227.22 \\
    iTransformer & 10.26\% & 239.83 \\
    TimesFM & 13.19\% & 308.73 \\
    Chronos1 & \underline{8.33\%} & \underline{194.61} \\
    Chronos2-small & \textbf{7.91\%} & \textbf{185.15} \\
    Chronos2-small (Uni-FT) & 9.57\% & 223.80 \\
    Chronos2-small (Multi-FT) & 8.85\% & 206.99 \\
    Time-LLM & 13.34\% & 312.68 \\
    VoT & 11.31\% & 264.57 \\
    ReasonCast (always-on semantics) & 9.59\% & 224.43 \\
    ReasonCast (agent-routed) & \textbf{7.91\%} & \textbf{185.15} \\
    \bottomrule
  \end{tabular}
\end{table}

Table~\ref{tab:stable-sales} isolates routing. Always-on semantics degrade the Chronos2-small backbone from 7.91\%/185.15 to 9.59\%/224.43. Agent-routed ReasonCast selects $\textsc{Skip}$ and recovers the backbone exactly. Event windows, by contrast, require at least $\textsc{Basic}$ reasoning.

\subsection{Dissecting Selective Semantic Intervention}
Table~\ref{tab:ablation} tests where, how strongly, and in what form semantics
enter the forecast. Its stable-sales column forces $\textsc{Basic}$ to isolate
fusion from routing; positive changes denote degradation.

\begin{table*}[!t]
  \centering
  \caption{Selective-fusion ablations (WMAPE/MAE). The stable-sales column forces $\textsc{Basic}$ to isolate fusion from routing. Gray values report changes from the full fusion mechanism as WMAPE percentage points / relative MAE (\%); positive values indicate degradation. Lower is better.}
  \label{tab:ablation}
  \setlength{\tabcolsep}{5.0pt}
  \renewcommand{\arraystretch}{1.18}
  \footnotesize
  \newcommand{\rcab}[4]{\shortstack{#1\%\,/\,#2\\[-1pt]
    \textcolor{gray}{\scriptsize $\Delta\;#3\,\mathrm{pp}\,/\,#4\%$}}}
  \newcommand{\rcref}[2]{\shortstack{#1\%\,/\,#2\\[-1pt]
    \textcolor{gray}{\scriptsize reference}}}
  \resizebox{\textwidth}{!}{%
  \begin{tabular}{@{}llcccc@{}}
    \toprule
    Design axis & Variant
      & Holiday Sens.
      & Mega-sale Sens.
      & \shortstack{Stable-sales\\(forced Basic)}
      & M5 \\
    \midrule
    \multicolumn{2}{l}{\textbf{Full fusion mechanism}}
      & \rcref{39.10}{1080.71}
      & \rcref{20.89}{589.94}
      & \rcref{9.59}{224.43}
      & \rcref{32.63}{4.00} \\
    \addlinespace[3pt]
    \textit{Interaction target} & w/o orthogonal decomposition--alignment
      & \rcab{40.04}{1155.23}{+0.94}{+6.90}
      & \rcab{21.14}{596.19}{+0.25}{+1.06}
      & \rcab{11.31}{264.57}{+1.72}{+17.89}
      & \rcab{32.64}{4.00}{+0.01}{+0.00} \\
    \addlinespace[2pt]
    \textit{Intervention strength} & w/o adaptive semantic gate
      & \rcab{39.76}{1112.21}{+0.66}{+2.91}
      & \rcab{21.54}{608.01}{+0.65}{+3.06}
      & \rcab{9.64}{225.53}{+0.05}{+0.49}
      & \rcab{32.64}{4.01}{+0.01}{+0.25} \\
    \addlinespace[2pt]
    \textit{Correction form} & w/o additive correction
      & \rcab{40.12}{1189.98}{+1.02}{+10.11}
      & \rcab{22.14}{622.09}{+1.25}{+5.45}
      & \rcab{8.85}{206.99}{-0.74}{-7.77}
      & \rcab{32.84}{4.03}{+0.21}{+0.75} \\
    & w/o multiplicative correction
      & \rcab{44.29}{1159.19}{+5.19}{+7.26}
      & \rcab{20.89}{589.95}{+0.00}{+0.00}
      & \rcab{9.59}{224.43}{+0.00}{+0.00}
      & \rcab{32.65}{4.01}{+0.02}{+0.25} \\
    \addlinespace[2pt]
    \textit{Reference} & Base fusion only
      & \rcab{44.30}{1229.05}{+5.20}{+13.73}
      & \rcab{21.50}{606.18}{+0.61}{+2.75}
      & \rcab{8.74}{204.43}{-0.85}{-8.91}
      & \rcab{33.03}{4.04}{+0.40}{+1.00} \\
    \bottomrule
  \end{tabular}%
  }
\end{table*}

Removing orthogonal decomposition--alignment degrades every internal slice, although its
M5 effect is negligible. Removing the adaptive gate also increases WMAPE in all
four slices, with much larger changes during event-sensitive periods; this
supports adaptive aggregate control but does not establish sample-level harm
rates. The correction paths are complementary: the additive path improves both
event regimes, whereas the multiplicative path is especially important for the
large holiday level shift. Under forced $\textsc{Basic}$, simplified fusion can
perform better on stable sales, confirming that unnecessary semantics still add
noise; discrete routing removes this failure mode through the exact no-text
path.

\subsection{Reasoner Post-Training: From Semantic Fidelity to Forecast Utility}
We evaluate cumulative reasoner stages with the same frozen fusion forecaster;
only the reasoning policy changes. Event WMAPE pools the absolute-error
numerator and demand denominator over holiday- and mega-sale-sensitive slices.
For seven-day instance $i$, let
$\ell_i^{m}=\sum_h|y_{i,h}-\hat y_{i,h}^{m}|/
(\sum_h|y_{i,h}|+\epsilon)$. We retain the sample-level negative transfer rate
$\mathrm{NTR}=N^{-1}\sum_i\mathbb{I}[\ell_i^{\mathrm{text}}>
\ell_i^{\mathrm{no\text{-}text}}]$, using the exact no-text path of the same
frozen forecaster.

\begin{table*}[!t]
  \centering
  \caption{Semantic fidelity and downstream forecast utility across cumulative
  reasoner post-training stages. Event WMAPE pools holiday-sensitive and
  mega-sale-sensitive observations. Negative Transfer Rate is the proportion
  of instances for which text-conditioned seven-day normalized absolute error
  exceeds the exact no-text forecast from the same frozen forecaster. Arrows
  indicate whether higher or lower values are better.}
  \label{tab:semantic-quality}
  \setlength{\tabcolsep}{3.0pt}
  \small
  \resizebox{\textwidth}{!}{%
  \begin{tabular}{lccccccc}
    \toprule
    Training stage
      & \shortstack{Parse\\Rate $\uparrow$}
      & \shortstack{Direction\\MAE $\downarrow$}
      & \shortstack{Shape\\Weighted-F1 $\uparrow$}
      & \shortstack{Amplitude\\log-MAE $\downarrow$}
      & \shortstack{Peak\\log-MAE $\downarrow$}
      & \shortstack{Event\\WMAPE $\downarrow$}
      & \shortstack{Negative Transfer\\Rate $\downarrow$} \\
    \midrule
    Original LLM
      & 91.36\% & 1.14 & 0.25 & 0.44 & 0.54 & 20.52\% & 23.59\% \\
    + Schema SFT
      & 100.00\% & 0.91 & 0.34 & 0.44 & 0.53 & 19.48\% & 16.20\% \\
    + Semantic-Field RL (GRPO)
      & 100.00\% & 0.85 & 0.33 & 0.39 & 0.52 & 19.16\% & 16.03\% \\
    + Forecast-Utility RL (GRPO)
      & 100.00\% & 0.83 & 0.35 & 0.38 & 0.53 & 18.96\% & 15.08\% \\
    \bottomrule
  \end{tabular}%
  }
\end{table*}

SFT establishes parseability and improves direction and shape; semantic-field RL further calibrates direction, amplitude, and peak. Forecast-utility RL targets Event WMAPE and NTR while preserving semantic fidelity, separating well-formed text from interventions that improve the final forecast.

\subsection{Entropy-Adaptive Semantic Intervention}
\label{sec:entropy-gate}
We isolate continuous gate control after a non-skip route. Within each forecast
window, item--window pairs are grouped into entropy tertiles and compared with
a global-$\alpha$ control that replaces the instance-wise gate with one shared
additive coefficient.

\begin{figure}[t]
  \centering
  \includegraphics[width=\columnwidth]{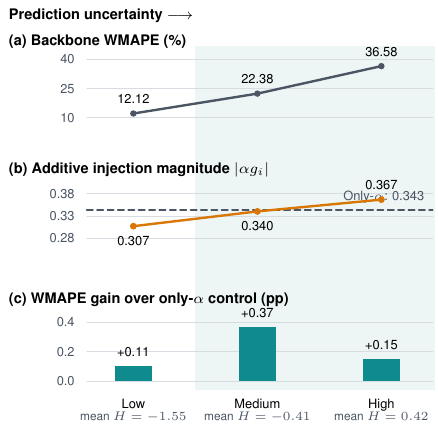}
  \caption{Entropy-adaptive semantic intervention across within-window forecast-entropy tertiles. (a) WMAPE of the no-text backbone; (b) mean additive-injection magnitude $|g_i|$, with the dashed line denoting the global-$\alpha$ control; and (c) WMAPE gain over that control in percentage points.}
  \Description{Three panels group forecasts into low-, medium-, and high-entropy tertiles, showing increasing no-text backbone WMAPE, increasing additive-injection magnitude, and positive WMAPE gains over a global-$\alpha$ control.}
  \label{fig:entropy-gate}
\end{figure}

Figure~\ref{fig:entropy-gate} links higher no-text entropy to stronger semantic
correction. Adaptive gating outperforms global $\alpha$ in every tier, with
larger gains at medium and high entropy, showing that intervention strength
tracks the reliability of the numerical forecast.

\paragraph{Semantic inversion stress test.} We invert all six structured semantic fields and the trend narrative while holding the time series, checkpoint, and inference configuration fixed. Incorrect semantics increase WMAPE in 13 of 15 windows, with mean and median increases of 2.67 and 0.76 percentage points and a worst-case increase of 25.33 points. Thus, the forecast responds to semantic content rather than merely to the presence of text; gating attenuates many corruptions but does not guarantee per-window immunity. Full window-level results appear in Table~\ref{tab:app-inversion}.

\section{Discussion}
Our results support interpreting event-enhanced forecasting as conditional intervention rather than uniform multimodal fusion. ReasonCast yields its clearest gains when future context supplies information that is difficult to recover from historical demand alone: relative to the strongest non-ReasonCast baselines, it reduces WMAPE by 7.8\% on holiday-sensitive categories, 5.6\% on mega-sale-sensitive categories, and 1.4\% on M5 event windows. In contrast, forcing semantic intervention during stable-sales periods increases WMAPE from 7.91\% to 9.59\%. This asymmetry is central to our formulation: semantic context has regime-dependent marginal utility and should modify a strong numerical forecaster only when it contributes information beyond the observed history.

The analyses further explain how ReasonCast realizes this principle. Discrete routing and continuous gating determine when semantic reasoning is invoked and how strongly it influences the forecast; the orthogonal decomposition determines where interaction occurs by restricting the semantic query interface to event-related temporal components; and the additive and multiplicative paths express different forms of correction. Removing orthogonal decomposition and alignment degrades all evaluated slices, while the multiplicative path is particularly important for holiday-driven level shifts. The additive path improves event-sensitive forecasts but can be harmful when intervention is unnecessarily forced on stable demand. Adaptive gating also outperforms a global-strength control across all forecast-entropy tiers, assigning greater influence to semantics when the numerical forecast is less reliable. Finally, semantic inversion increases WMAPE in 13 of 15 windows, confirming that the forecast responds to semantic content rather than merely to the presence of text. These results support forecast-conditioned semantic interaction, although they do not imply that the learned event subspace is a causally identifiable factor.

The stable-sales analysis clarifies the distinct roles of routing and fusion. In Table~\ref{tab:ablation}, the stable-sales column deliberately overrides the routing policy and forces every instance to take the \textsc{Basic} route. It therefore diagnoses how the fusion mechanism behaves under unnecessary intervention rather than reporting end-to-end ReasonCast performance. In deployment, the agent may instead choose \textsc{Skip}, disable both correction paths, and exactly recover the numerical backbone. The evidence thus supports two narrower claims: the fusion components improve forecasting when semantic intervention is warranted, while the exact-identity skip path prevents the model from systematically modifying the backbone when no intervention is selected. It does not imply that the fusion gate alone guarantees the absence of sample-level negative transfer.

The post-training results highlight a related distinction between semantic fidelity and forecast utility. Schema SFT establishes parseable and internally consistent outputs, and semantic-field RL improves forecast-critical judgments such as direction, amplitude, and peak intensity. Forecast-utility RL then evaluates valid candidates through a frozen forecaster, incorporating marginal predictive improvement, potential harm, and tool cost into the reasoning objective. This separation is important because well-formed event descriptions are not necessarily useful interventions.

Several broader challenges remain for deploying event-aware forecasting systems. Event effects are intrinsically non-stationary across products, regions, and seasons, requiring semantic interpretation and forecast intervention to be recalibrated under distribution shift. External signals also vary in timeliness, completeness, and reliability, making evidence validation and uncertainty-aware reasoning essential. At production scale, tool-augmented reasoning introduces a fundamental trade-off among predictive accuracy, latency, and computational cost, while sparse counterfactual observations make it difficult to separate forecast-relevant associations from causal event effects. A specific limitation of the current study is that it does not yet provide a complete accounting of tool-call frequency, end-to-end latency, or realized inference cost. Future work may combine online adaptation, calibrated abstention, source-reliability modeling, and resource-aware routing to make semantic intervention more robust and operationally efficient.

\section{Conclusion}
We presented ReasonCast, an agentic framework that treats event semantics as a routed intervention rather than an always-on forecasting input. The agent selects among no intervention, low-cost structured reasoning, and tool-augmented reasoning, while continuous gating controls intervention strength after a non-skip decision. Schema SFT, semantic-field RL, and forecast-utility RL progressively align structured reasoning with downstream predictive value. The results show that forecast-aligned routing can exploit contextual signals during event-driven shifts while exactly preserving the numerical backbone when semantics are unnecessary, thereby avoiding systematic negative transfer in stable-sales periods.

\bibliographystyle{ACM-Reference-Format}
\bibliography{sample-base}

\clearpage
\onecolumn
\appendix
\section{Detailed Window-Level Results}
\label{app:detailed-results}

This appendix reports the unaggregated seven-day forecast windows underlying
the main-text averages. Each cell contains WMAPE and MAE for exactly the same
items, forecast origin, and horizon; no window-specific model selection is
performed. For compactness, Chronos2-small is abbreviated as C2-small and the
time-series backbone as TS backb.

\begin{minipage}[t]{0.485\textwidth}
\subsection{Commerce Dataset}

Tables~\ref{tab:app-commerce-a} and~\ref{tab:app-commerce-b} contain the 15
commerce windows in chronological blocks: three mega-sale windows, ten
holiday windows, and two stable-sales windows. The split across two tables is
purely presentational; the rows and evaluation protocol are identical.

The hardest cases are the two Spring Festival--Valentine windows beginning on
February 12 and February 15. On February 12, ReasonCast reduces WMAPE from
108.75\% for the strongest non-ReasonCast comparator in its table to 83.85\%;
on February 15 it further reduces the best comparator from 28.77\% to
24.53\%. These windows combine a sharp holiday regime change with overlapping
event semantics, precisely the setting in which history-only extrapolation is
least reliable.

The stable-sales rows provide an important counterpoint. ReasonCast is not the
best model on January 13 or March 23: several purely numerical models attain
lower error. The aggregate benefit therefore does not come from uniformly
overriding the backbone. Instead, the detailed rows support the intended
division of labor: semantic intervention is most useful when future events
add information absent from the observed history, whereas the numerical
backbone remains highly competitive when demand is stable.
\end{minipage}\hfill
\begin{minipage}[t]{0.485\textwidth}

\subsection{M5 Dataset}

Tables~\ref{tab:app-m5-a} and~\ref{tab:app-m5-b} report all 14 M5 event
windows. Each holiday is evaluated at two forecast origins, which prevents a
single favorable alignment between the event day and horizon from determining
the aggregate result.

The public benchmark has a different error profile from the commerce data.
Absolute differences are smaller, but ReasonCast remains consistently
competitive across Halloween, New Year, Super Bowl, Easter, and Mother's Day
origins. It does not dominate every origin: the time-series backbone is
slightly better on the first Christmas origin, and the covariate baseline is
slightly better on the first Thanksgiving origin. Reporting both origins
therefore exposes where semantic context helps and where a strong numerical
model already captures most of the predictable event pattern.
\end{minipage}

\begin{table}[H]
  \centering
  \caption{Commerce window-level results, baseline group A. Each method
  occupies a WMAPE (\%) and MAE pair. Rows 1--3 are mega-sale windows,
  rows 4--13 are holiday windows, and rows 14--15 are stable-sales windows.}
  \label{tab:app-commerce-a}
  \scriptsize
  \setlength{\tabcolsep}{2.2pt}
  \renewcommand{\arraystretch}{1.08}
  \resizebox{\textwidth}{!}{%
  \begin{tabular}{lrr@{\hspace{4pt}}rr@{\hspace{4pt}}rr@{\hspace{4pt}}rr@{\hspace{4pt}}rr@{\hspace{4pt}}rr}
    \toprule
    & \multicolumn{2}{c}{SNaive}
    & \multicolumn{2}{c}{LightGBM}
    & \multicolumn{2}{c}{PatchTST}
    & \multicolumn{2}{c}{iTransformer}
    & \multicolumn{2}{c}{TimesFM}
    & \multicolumn{2}{c}{C2-small} \\
    \cmidrule(lr){2-3}\cmidrule(lr){4-5}\cmidrule(lr){6-7}
    \cmidrule(lr){8-9}\cmidrule(lr){10-11}\cmidrule(lr){12-13}
    Window & WMAPE & MAE & WMAPE & MAE & WMAPE & MAE
      & WMAPE & MAE & WMAPE & MAE & WMAPE & MAE \\
    \midrule
    2025-11-11 & 19.61\% & 547.89 & 18.44\% & 515.27 & 19.68\% & 550.02 & 19.63\% & 548.61 & 23.87\% & 667.14 & 18.03\% & 509.35 \\
    2025-12-08 & 15.55\% & 392.93 & 17.21\% & 434.92 & 10.72\% & 270.99 & 12.40\% & 313.53 & 15.12\% & 382.17 & 15.94\% & 406.74 \\
    2025-12-11 & 14.51\% & 363.26 & 13.04\% & 326.56 & 9.84\% & 246.23 & 11.43\% & 286.05 & 13.89\% & 347.66 & 9.94\% & 251.24 \\
    \addlinespace
    2026-01-01 & 18.68\% & 446.77 & 17.61\% & 421.21 & 18.82\% & 450.14 & 18.36\% & 439.06 & 19.72\% & 471.68 & 29.83\% & 719.76 \\
    2026-02-12 & 210.75\% & 1614.50 & 169.72\% & 1300.13 & 175.83\% & 1347.00 & 159.53\% & 1222.10 & 186.35\% & 1427.55 & 138.84\% & 1070.39 \\
    2026-02-15 & 131.99\% & 1077.18 & 54.45\% & 444.35 & 107.73\% & 879.20 & 53.02\% & 432.70 & 67.55\% & 551.23 & 48.13\% & 395.08 \\
    2026-03-02 & 16.67\% & 397.63 & 15.35\% & 366.01 & 12.78\% & 304.67 & 14.74\% & 351.40 & 15.92\% & 379.54 & 15.11\% & 362.31 \\
    2026-03-05 & 14.25\% & 336.66 & 12.75\% & 301.27 & 12.37\% & 292.26 & 14.06\% & 332.20 & 17.03\% & 402.42 & 12.23\% & 290.56 \\
    2026-04-28 & 20.43\% & 388.59 & 19.54\% & 371.61 & 20.23\% & 384.60 & 21.66\% & 411.86 & 29.08\% & 553.03 & 20.86\% & 396.73 \\
    2026-05-04 & 23.19\% & 534.31 & 16.59\% & 382.22 & 16.47\% & 379.51 & 17.25\% & 397.43 & 19.30\% & 444.67 & 19.84\% & 457.16 \\
    2026-05-10 & 16.17\% & 391.96 & 14.31\% & 346.96 & 13.93\% & 337.68 & 14.39\% & 348.90 & 14.68\% & 355.83 & 11.73\% & 284.37 \\
    2026-05-16 & 11.51\% & 280.86 & 10.82\% & 263.98 & 9.52\% & 232.32 & 12.55\% & 306.24 & 10.80\% & 263.47 & 8.03\% & 195.89 \\
    2026-05-19 & 10.23\% & 250.58 & 11.14\% & 273.05 & 10.10\% & 247.35 & 10.80\% & 264.61 & 13.27\% & 325.10 & 8.22\% & 201.46 \\
    \addlinespace
    2026-01-13 & 10.02\% & 236.73 & 10.88\% & 257.11 & 8.46\% & 199.75 & 9.10\% & 214.96 & 13.12\% & 309.89 & 7.49\% & 176.95 \\
    2026-03-23 & 11.77\% & 272.90 & 11.33\% & 262.64 & 10.98\% & 254.70 & 11.41\% & 264.70 & 13.26\% & 307.57 & 8.34\% & 193.35 \\
    \bottomrule
  \end{tabular}}
\end{table}

\begin{table}[H]
  \centering
  \caption{Commerce window-level results, baseline group B and ReasonCast.
  Each method occupies a WMAPE (\%) and MAE pair.}
  \label{tab:app-commerce-b}
  \scriptsize
  \setlength{\tabcolsep}{2.2pt}
  \renewcommand{\arraystretch}{1.08}
  \resizebox{\textwidth}{!}{%
  \begin{tabular}{lrr@{\hspace{4pt}}rr@{\hspace{4pt}}rr@{\hspace{4pt}}rr@{\hspace{4pt}}rr@{\hspace{4pt}}rr}
    \toprule
    & \multicolumn{2}{c}{TS backb.}
    & \multicolumn{2}{c}{+ covariates}
    & \multicolumn{2}{c}{Chronos}
    & \multicolumn{2}{c}{Time-LLM}
    & \multicolumn{2}{c}{VoT}
    & \multicolumn{2}{c}{\textbf{ReasonCast}} \\
    \cmidrule(lr){2-3}\cmidrule(lr){4-5}\cmidrule(lr){6-7}
    \cmidrule(lr){8-9}\cmidrule(lr){10-11}\cmidrule(lr){12-13}
    Window & WMAPE & MAE & WMAPE & MAE & WMAPE & MAE
      & WMAPE & MAE & WMAPE & MAE & WMAPE & MAE \\
    \midrule
    2025-11-11 & 18.04\% & 504.26 & 17.74\% & 495.69 & 17.67\% & 493.82 & 20.11\% & 561.93 & 22.86\% & 638.72 & 17.18\% & 480.21 \\
    2025-12-08 & 13.31\% & 336.44 & 11.61\% & 293.52 & 16.84\% & 425.55 & 16.42\% & 415.09 & 15.13\% & 382.42 & 11.25\% & 284.27 \\
    2025-12-11 & 12.30\% & 307.93 & 10.69\% & 267.51 & 10.20\% & 255.44 & 14.38\% & 359.93 & 12.51\% & 313.13 & 10.17\% & 254.56 \\
    \addlinespace
    2026-01-01 & 24.92\% & 596.10 & 27.36\% & 654.40 & 27.72\% & 663.06 & 15.69\% & 375.23 & 19.89\% & 475.70 & 22.29\% & 533.21 \\
    2026-02-12 & 131.66\% & 1008.59 & 116.26\% & 890.61 & 141.52\% & 1084.17 & 210.34\% & 1611.33 & 108.75\% & 833.09 & 83.85\% & 642.36 \\
    2026-02-15 & 35.95\% & 293.38 & 28.77\% & 234.77 & 33.26\% & 271.42 & 136.70\% & 1115.56 & 42.28\% & 345.02 & 24.53\% & 200.21 \\
    2026-03-02 & 11.09\% & 264.51 & 10.82\% & 258.05 & 14.24\% & 339.45 & 16.67\% & 397.43 & 16.19\% & 386.05 & 10.65\% & 253.93 \\
    2026-03-05 & 9.03\% & 213.45 & 8.88\% & 209.76 & 10.79\% & 254.92 & 15.62\% & 369.00 & 12.63\% & 298.44 & 8.78\% & 207.43 \\
    2026-04-28 & 14.44\% & 274.63 & 13.13\% & 249.78 & 20.35\% & 386.90 & 20.34\% & 386.70 & 17.95\% & 341.31 & 12.52\% & 238.09 \\
    2026-05-04 & 17.88\% & 411.98 & 16.83\% & 387.78 & 20.32\% & 468.08 & 19.92\% & 458.93 & 18.32\% & 422.06 & 16.34\% & 376.43 \\
    2026-05-10 & 15.42\% & 373.72 & 14.70\% & 356.27 & 13.05\% & 316.35 & 13.34\% & 323.52 & 16.22\% & 393.13 & 13.76\% & 333.46 \\
    2026-05-16 & 9.78\% & 238.51 & 10.01\% & 244.12 & 7.56\% & 184.29 & 10.45\% & 254.90 & 11.48\% & 280.11 & 9.07\% & 221.16 \\
    2026-05-19 & 8.69\% & 213.02 & 9.10\% & 223.05 & 8.23\% & 201.68 & 11.20\% & 274.46 & 9.58\% & 234.69 & 8.82\% & 216.14 \\
    \addlinespace
    2026-01-13 & 8.43\% & 199.07 & 8.43\% & 199.24 & 6.91\% & 163.30 & 15.35\% & 362.56 & 10.59\% & 250.25 & 9.60\% & 226.74 \\
    2026-03-23 & 10.72\% & 248.52 & 9.26\% & 214.74 & 9.74\% & 225.93 & 11.33\% & 262.80 & 12.03\% & 278.88 & 9.58\% & 222.12 \\
    \bottomrule
  \end{tabular}}
\end{table}

\begin{table}[H]
  \centering
  \caption{M5 window-level results, baseline group A. Each method occupies a
  WMAPE (\%) and MAE pair.}
  \label{tab:app-m5-a}
  \scriptsize
  \setlength{\tabcolsep}{2.2pt}
  \renewcommand{\arraystretch}{1.08}
  \resizebox{\textwidth}{!}{%
  \begin{tabular}{lrr@{\hspace{4pt}}rr@{\hspace{4pt}}rr@{\hspace{4pt}}rr@{\hspace{4pt}}rr@{\hspace{4pt}}rr}
    \toprule
    & \multicolumn{2}{c}{SNaive}
    & \multicolumn{2}{c}{LightGBM}
    & \multicolumn{2}{c}{PatchTST}
    & \multicolumn{2}{c}{iTransformer}
    & \multicolumn{2}{c}{TimesFM}
    & \multicolumn{2}{c}{C2-small} \\
    \cmidrule(lr){2-3}\cmidrule(lr){4-5}\cmidrule(lr){6-7}
    \cmidrule(lr){8-9}\cmidrule(lr){10-11}\cmidrule(lr){12-13}
    Event and origin & WMAPE & MAE & WMAPE & MAE & WMAPE & MAE
      & WMAPE & MAE & WMAPE & MAE & WMAPE & MAE \\
    \midrule
    Halloween 10-25 & 37.20\% & 4.55 & 29.25\% & 3.58 & 45.65\% & 5.59 & 29.45\% & 3.61 & 29.84\% & 3.65 & 29.29\% & 3.59 \\
    Halloween 10-31 & 38.36\% & 4.86 & 30.46\% & 3.86 & 44.79\% & 5.68 & 30.71\% & 3.89 & 30.17\% & 3.82 & 29.38\% & 3.72 \\
    Thanksgiving 11-21 & 46.84\% & 5.45 & 35.41\% & 4.12 & 53.78\% & 6.26 & 36.71\% & 4.27 & 37.91\% & 4.41 & 36.64\% & 4.27 \\
    Thanksgiving 11-24 & 50.78\% & 5.46 & 37.95\% & 4.08 & 56.76\% & 6.11 & 41.32\% & 4.44 & 41.86\% & 4.50 & 41.30\% & 4.44 \\
    Christmas 12-21 & 59.59\% & 6.17 & 49.74\% & 5.15 & 68.18\% & 7.06 & 49.61\% & 5.13 & 50.30\% & 5.21 & 49.69\% & 5.14 \\
    Christmas 12-24 & 64.00\% & 6.18 & 51.91\% & 5.01 & 71.71\% & 6.92 & 52.24\% & 5.04 & 53.95\% & 5.21 & 54.07\% & 5.22 \\
    NewYear 12-27 & 52.48\% & 6.19 & 35.05\% & 4.14 & 49.38\% & 5.83 & 34.11\% & 4.02 & 33.76\% & 3.98 & 33.91\% & 4.00 \\
    NewYear 12-30 & 49.21\% & 6.25 & 33.68\% & 4.28 & 46.54\% & 5.91 & 32.62\% & 4.14 & 32.30\% & 4.10 & 32.79\% & 4.16 \\
    SuperBowl 02-01 & 37.92\% & 5.22 & 29.63\% & 4.07 & 42.02\% & 5.78 & 30.21\% & 4.15 & 29.82\% & 4.10 & 30.04\% & 4.13 \\
    SuperBowl 02-07 & 40.13\% & 5.72 & 30.98\% & 4.41 & 41.19\% & 5.87 & 31.03\% & 4.42 & 30.65\% & 4.37 & 31.17\% & 4.44 \\
    Easter 03-23 & 37.29\% & 4.97 & 29.70\% & 3.96 & 41.53\% & 5.54 & 29.23\% & 3.90 & 29.57\% & 3.94 & 29.89\% & 3.98 \\
    Easter 03-26 & 37.90\% & 5.02 & 28.55\% & 3.78 & 41.63\% & 5.51 & 29.50\% & 3.91 & 30.03\% & 3.98 & 30.20\% & 4.00 \\
    Mothers day 05-04 & 36.55\% & 5.31 & 27.77\% & 4.03 & 36.62\% & 5.32 & 27.95\% & 4.06 & 28.01\% & 4.07 & 27.26\% & 3.96 \\
    Mothers day 05-07 & 35.45\% & 5.19 & 26.71\% & 3.91 & 36.58\% & 5.35 & 26.82\% & 3.93 & 26.65\% & 3.90 & 25.95\% & 3.80 \\
    \bottomrule
  \end{tabular}}
\end{table}

\begin{table}[H]
  \centering
  \caption{M5 window-level results, baseline group B and ReasonCast. Each
  method occupies a WMAPE (\%) and MAE pair.}
  \label{tab:app-m5-b}
  \scriptsize
  \setlength{\tabcolsep}{2.2pt}
  \renewcommand{\arraystretch}{1.08}
  \resizebox{\textwidth}{!}{%
  \begin{tabular}{lrr@{\hspace{4pt}}rr@{\hspace{4pt}}rr@{\hspace{4pt}}rr@{\hspace{4pt}}rr@{\hspace{4pt}}rr}
    \toprule
    & \multicolumn{2}{c}{TS backb.}
    & \multicolumn{2}{c}{+ covariates}
    & \multicolumn{2}{c}{Chronos}
    & \multicolumn{2}{c}{Time-LLM}
    & \multicolumn{2}{c}{VoT}
    & \multicolumn{2}{c}{\textbf{ReasonCast}} \\
    \cmidrule(lr){2-3}\cmidrule(lr){4-5}\cmidrule(lr){6-7}
    \cmidrule(lr){8-9}\cmidrule(lr){10-11}\cmidrule(lr){12-13}
    Event and origin & WMAPE & MAE & WMAPE & MAE & WMAPE & MAE
      & WMAPE & MAE & WMAPE & MAE & WMAPE & MAE \\
    \midrule
    Halloween 10-25 & 29.59\% & 3.62 & 29.24\% & 3.58 & 30.50\% & 3.73 & 35.20\% & 4.31 & 33.53\% & 4.11 & 28.83\% & 3.53 \\
    Halloween 10-31 & 30.52\% & 3.87 & 29.76\% & 3.77 & 30.55\% & 3.87 & 35.21\% & 4.46 & 35.73\% & 4.53 & 29.52\% & 3.74 \\
    Thanksgiving 11-21 & 33.77\% & 3.93 & 33.46\% & 3.89 & 37.69\% & 4.39 & 42.81\% & 4.98 & 38.81\% & 4.52 & 33.81\% & 3.94 \\
    Thanksgiving 11-24 & 35.85\% & 3.86 & 36.40\% & 3.92 & 41.53\% & 4.47 & 45.59\% & 4.90 & 39.83\% & 4.28 & 35.71\% & 3.84 \\
    Christmas 12-21 & 47.10\% & 4.87 & 47.72\% & 4.94 & 50.25\% & 5.20 & 54.21\% & 5.61 & 53.72\% & 5.56 & 48.34\% & 5.00 \\
    Christmas 12-24 & 48.43\% & 4.67 & 48.65\% & 4.69 & 54.20\% & 5.23 & 57.83\% & 5.58 & 52.18\% & 5.04 & 48.46\% & 4.68 \\
    NewYear 12-27 & 33.74\% & 3.98 & 33.14\% & 3.91 & 35.64\% & 4.21 & 37.86\% & 4.47 & 37.64\% & 4.44 & 32.49\% & 3.83 \\
    NewYear 12-30 & 32.65\% & 4.14 & 32.59\% & 4.14 & 34.97\% & 4.44 & 38.05\% & 4.83 & 39.46\% & 5.01 & 31.53\% & 4.00 \\
    SuperBowl 02-01 & 29.33\% & 4.03 & 29.35\% & 4.04 & 31.00\% & 4.26 & 34.26\% & 4.71 & 35.39\% & 4.87 & 29.18\% & 4.01 \\
    SuperBowl 02-07 & 31.36\% & 4.47 & 30.78\% & 4.38 & 31.36\% & 4.47 & 33.89\% & 4.83 & 36.16\% & 5.15 & 30.69\% & 4.37 \\
    Easter 03-23 & 29.87\% & 3.98 & 30.21\% & 4.03 & 30.40\% & 4.05 & 34.76\% & 4.63 & 33.63\% & 4.48 & 29.28\% & 3.90 \\
    Easter 03-26 & 29.18\% & 3.87 & 29.07\% & 3.85 & 30.63\% & 4.06 & 34.19\% & 4.53 & 34.63\% & 4.59 & 28.62\% & 3.79 \\
    Mothers day 05-04 & 27.36\% & 3.97 & 27.08\% & 3.93 & 28.64\% & 4.16 & 31.35\% & 4.55 & 31.25\% & 4.54 & 26.87\% & 3.90 \\
    Mothers day 05-07 & 26.19\% & 3.83 & 25.98\% & 3.80 & 27.96\% & 4.09 & 30.69\% & 4.49 & 30.72\% & 4.50 & 25.54\% & 3.74 \\
    \bottomrule
  \end{tabular}}
\end{table}

\subsection{Semantic Corruption Stress Test}

Table~\ref{tab:app-inversion} gives the per-window results for the controlled
semantic inversion test. We invert the six control fields and the final trend
narrative while holding the time-series input, checkpoint, and inference
configuration fixed. Positive $\Delta$WMAPE denotes degradation. Incorrect
semantics increase WMAPE in 13 of 15 windows, by 2.67 percentage points on
average.

The distribution is strongly right-skewed. The median increase is 0.76
percentage points, while the February 12 Spring Festival--Valentine window
increases by 25.33 points. Thus, a single aggregate mean would understate the
typical harm while obscuring the severe failure mode on a highly
event-sensitive window. The two small negative deltas are also informative:
gating attenuates many incorrect interventions, but it cannot guarantee that
every corrupted description worsens every finite test slice.

\begin{table}[H]
  \centering
  \caption{Semantic inversion by window. $\Delta$ is inverted minus correct
  WMAPE in percentage points.}
  \label{tab:app-inversion}
  \small
  \setlength{\tabcolsep}{8pt}
  \renewcommand{\arraystretch}{1.12}
  \begin{tabular}{llrrrrr}
    \toprule
    Event & Window & \shortstack{Correct\\WMAPE}
      & \shortstack{Inverted\\WMAPE} & $\Delta$
      & \shortstack{Correct\\MAE} & \shortstack{Inverted\\MAE} \\
    \midrule
    520 & 2026-05-16 & 7.38\% & 7.77\% & 0.39 & 252.36 & 265.70 \\
    520 & 2026-05-19 & 5.78\% & 6.47\% & 0.69 & 218.79 & 244.81 \\
    Double 11 & 2025-11-05 & 13.27\% & 14.03\% & 0.76 & 154.35 & 163.25 \\
    Double 11 & 2025-11-08 & 17.54\% & 19.15\% & 1.61 & 959.36 & 1047.42 \\
    Double 11 & 2025-11-11 & 13.76\% & 13.62\% & -0.15 & 618.88 & 612.33 \\
    Double 12 & 2025-12-08 & 9.89\% & 11.09\% & 1.19 & 163.66 & 183.39 \\
    Double 12 & 2025-12-11 & 9.77\% & 12.15\% & 2.39 & 184.16 & 229.16 \\
    Labor Day & 2026-04-28 & 7.61\% & 8.74\% & 1.14 & 175.76 & 202.01 \\
    Labor Day & 2026-05-04 & 13.98\% & 14.71\% & 0.74 & 457.55 & 481.63 \\
    Lantern & 2026-03-02 & 8.42\% & 9.61\% & 1.19 & 225.67 & 257.55 \\
    Mother's Day & 2026-05-10 & 12.35\% & 12.89\% & 0.53 & 403.99 & 421.43 \\
    New Year & 2026-01-01 & 15.13\% & 15.09\% & -0.04 & 403.91 & 402.86 \\
    Spring--Valentine & 2026-02-12 & 79.37\% & 104.69\% & 25.33 & 943.54 & 1244.66 \\
    Spring--Valentine & 2026-02-15 & 21.21\% & 24.95\% & 3.74 & 286.61 & 337.10 \\
    Women's Day & 2026-03-05 & 7.66\% & 8.16\% & 0.50 & 206.44 & 219.89 \\
    \bottomrule
  \end{tabular}
\end{table}

\clearpage
\twocolumn
\section{Agent Policy, Semantic Interpretation, and Optimization Details}
\label{app:reasoner-details}

\subsection{Agent Policy and Basic Semantic Interpreter Prompt}

We explicitly separate the \emph{agent policy} from the \emph{semantic interpreter}. Given the event context, the no-text forecast, and its predictive entropy, the policy selects $\textsc{Skip}$, $\textsc{Basic}$, or $\textsc{Tool}$. $\textsc{Skip}$ terminates semantic processing and preserves the numerical backbone. $\textsc{Basic}$ invokes the semantic interpreter with the available context. $\textsc{Tool}$ first augments that context with retrieved event evidence or temporal statistics and then invokes the same interpreter. Holidays and mega-sales disallow $\textsc{Skip}$; other instances may abstain when the backbone is sufficiently reliable.

The fixed procedure below is therefore not the agent policy. We retain it as the \emph{Basic semantic interpreter prompt}, invoked only after a non-skip policy decision. The $\textsc{Basic}$ route supplies the original context, whereas the $\textsc{Tool}$ route supplies tool-augmented context. All scale judgments are expressed relative to the most recent seven-day mean.

\begin{figure}[H]
  \centering
  \fbox{%
  \begin{minipage}{0.94\columnwidth}
    \footnotesize
    \begin{tabular}{@{}r p{0.73\columnwidth}@{}}
      \textsc{Input:} &
      item hierarchy; forecast dates; event calendar and event positions;
      recent seven-day demand, mean, and slope; no-text forecast and entropy;
      current level $C$; aligned historical mean $A$ and peak
      $A_{\mathrm{peak}}$ when available; platform activities; and any
      policy-approved tool observations. \\[2pt]
      1: & Validate the reference scale. If aligned history exists, compute
      $A/C$ for the seven-day mean effect and $A_{\mathrm{peak}}/C$ for the
      single-day peak; otherwise activate the no-YoY branch. \\
      2: & Assess item--event relevance from historical evidence, item
      knowledge, recent slope, and event position. Do not infer relevance
      solely from the presence of an event. \\
      3: & Select the dominant event when multiple events or promotions
      overlap, and record why weaker candidates are excluded. \\
      4: & Assign demand direction and temporal shape, explicitly separating
      the seven-day mean from a concentrated single-day peak. \\
      5: & Calibrate amplitude and peak multipliers to the recent seven-day
      reference, using conservative priors when aligned history is absent. \\
      6: & Check cross-field consistency: relevance, direction, shape,
      amplitude, peak, and peak timing must describe the same trajectory. \\
      \textsc{Return:} &
      six structured control fields plus one concise trend narrative. Never
      directly generate the numerical demand forecast. \\
    \end{tabular}
  \end{minipage}}
  \caption{Basic semantic interpreter prompt used after a non-skip policy decision. The tool route augments its input evidence but does not change the output schema.}
  \Description{A fixed semantic interpreter validates the reference scale, assesses event relevance, selects a dominant event, predicts direction and temporal shape, calibrates magnitude and peak effects, checks consistency, and returns structured semantic fields.}
  \label{fig:app-basic-interpreter}
\end{figure}

This separation also defines the roles of the three post-training stages. Schema SFT and semantic-field RL train the semantic interpreter to produce valid, consistent, and calibrated forecast-specific fields. Forecast-utility RL trains the agent policy---including route selection, tool use, and the final intervention decision---against realized downstream utility from the frozen forecaster, with explicit penalties for negative transfer and tool cost.

\subsection{Structured Fields and Representative Rationale}

Every final policy response begins with one machine-readable routing action. A $\textsc{Skip}$ response terminates there and activates the no-text path. $\textsc{Basic}$ and $\textsc{Tool}$ responses additionally contain six machine-readable control fields and a short natural-language trend narrative. Tool calls and observations belong to the intermediate policy trajectory rather than the final output schema. The narrative is encoded together with the fields, but it is not counted as an additional control label because it has no closed label set.

\begin{table}[H]
  \centering
  \caption{ReasonCast reasoner output schema.}
  \label{tab:app-output-schema}
  \footnotesize
  \setlength{\tabcolsep}{3.5pt}
  \begin{tabular}{p{0.20\columnwidth}p{0.72\columnwidth}}
    \toprule
    Field & Value space and forecasting role \\
    \midrule
    Intervention action
      & Skip, basic, or tool. Controls whether semantic generation is bypassed,
        uses available context, or invokes additional evidence/statistics. \\
    Relevance
      & Related / unrelated. Determines whether the context is a plausible
        intervention rather than incidental calendar text. \\
    Dominant event
      & Event name / none. Resolves overlapping holidays and promotions to a
        primary driver. \\
    Direction
      & Large decrease, small decrease, flat, small increase, or large
        increase; describes the seven-day mean relative to the recent mean. \\
    Temporal shape
      & Increasing, decreasing, rise--fall, fall--rise, or flat; specifies the
        within-horizon trajectory and turning behavior. \\
    Amplitude
      & Positive scalar; ratio of the predicted seven-day mean to the recent
        seven-day mean. \\
    Peak
      & Positive scalar; ratio of the largest predicted day to the recent
        seven-day mean. \\
    Trend narrative
      & At most 80 Chinese characters; states event timing, turning point, and
        coarse magnitude for semantic encoding. \\
    \bottomrule
  \end{tabular}
\end{table}

Table~\ref{tab:app-rationale-example} shows a representative teacher example.
The stored compressed rationale is Chinese; an English translation is shown
for readability. It is a concise forecasting rationale, not a raw
token-by-token trace of the teacher's internal generation.

\begin{table}[!t]
  \centering
  \caption{Representative compressed rationale and structured output for a
  laptop forecast from March 7 to March 13, with Women's Day on horizon day 2.}
  \label{tab:app-rationale-example}
  \footnotesize
  \setlength{\tabcolsep}{4.0pt}
  \begin{tabular}{p{0.93\columnwidth}}
    \toprule
    Compressed rationale (English translation) \\
    \midrule
    The window covers the March 8 platform campaign on day 2. Laptops are not
    a core Women's Day gift category, but they respond to platform-wide
    promotions. Recent demand is already elevated, so the additional effect
    should be moderate rather than comparable with Double 11 or 618. We
    therefore expect a short peak of about $1.4\times$ on March 8 followed by
    a return toward the recent level, producing a seven-day mean of about
    $1.2\times$. \\
    \midrule
    Structured target \\
    \midrule
    \textbf{Relevance:} related\newline
    \textbf{Dominant event:} Women's Day\newline
    \textbf{Direction:} small increase\newline
    \textbf{Shape:} rise--fall\newline
    \textbf{Amplitude:} 1.2\newline
    \textbf{Peak:} 1.4\newline
    \textbf{Trend:} promotion-driven peak on March 8, followed by a return
    toward the recent level. \\
    \bottomrule
  \end{tabular}
\end{table}

\subsection{Teacher Distillation and Rationale Compression}

We construct the initial SFT pool by stratifying 40,000 item--window examples
across upward (40\%), downward (22\%), flat (33\%), and noisy (5\%) demand
regimes. Claude Opus 4.6 serves as the teacher and receives only the Basic semantic interpreter prompt, without agent routing or tool interaction. This stage deliberately supervises structured semantic construction rather than the agent policy. Of 11,627 successful teacher generations, deterministic filters on direction, amplitude, and peak consistency retain 7,312 candidates.

The teacher rationales are substantially longer than needed by the student
model. We therefore translate the English rationales into Chinese and compress
them to a single 200--300-character paragraph while preserving the original
decision order: item and window, event position, recent baseline and $C$,
historical $A$ and $A_{\mathrm{peak}}$ or the no-YoY branch, item--event
relevance, dominant-event selection, direction and shape, magnitude and peak,
and any necessary recalibration. Numerical evidence may only be copied from
the prompt or teacher response. The six structured labels are held fixed
during compression. After compression validation, deduplication, and
rebalancing, the final SFT set contains 4,590 examples.

This data transformation shortens the student rationale without discarding
the forecasting decision pattern. As shown in
Table~\ref{tab:semantic-quality}, mean rationale length decreases from 568.99
to 206.61 tokens after SFT and to 202.77 tokens after semantic-field RL, a
64.36\% reduction relative to the original LLM. Direction, amplitude, and peak
errors simultaneously decline. Because Table~\ref{tab:semantic-quality}
reports cumulative post-training stages rather than an isolated compression
ablation, we interpret this as an end-to-end distillation result rather than
attributing the quality gain to compression alone.

\subsection{SFT, Semantic-Field RL, and Forecast-Utility RL Configuration}

Tables~\ref{tab:app-training-config} and~\ref{tab:app-stage2-config} record the
main optimization settings. Both RL stages use group-relative policy optimization (GRPO). Semantic-field RL improves the forecast-critical fields; forecast-utility RL first applies a semantic validity gate and then evaluates only gate-passing candidates with the frozen fusion model.

\newpage
\begin{table}[H]
  \centering
  \caption{SFT and semantic-field RL configuration.}
  \label{tab:app-training-config}
  \footnotesize
  \setlength{\tabcolsep}{3.5pt}
  \begin{tabular}{p{0.26\columnwidth}p{0.66\columnwidth}}
    \toprule
    Component & Configuration \\
    \midrule
    Student model
      & Qwen3-32B~\cite{yang2025qwen3} for SFT and both GRPO stages. \\
    SFT data
      & 4,590 compressed teacher examples; 2\% validation split. \\
    SFT adapter
      & LoRA on all linear layers, rank 16, scale 32, dropout 0.05. \\
    SFT optimization
      & Three epochs; learning rate $1\times10^{-4}$; cosine schedule;
        warmup ratio 0.03; BF16; effective batch size 32
        (eight workers, batch size 1, four accumulation steps). \\
    Semantic-field adapter
      & New LoRA initialized on the SFT-merged model; rank 8, scale 16,
        dropout 0; all linear layers. \\
    Semantic-field sampling
      & Group size 8; temperature 0.8; top-$p$ 0.95; top-$k$ 50; maximum
        prompt/completion lengths 2816/512 tokens. \\
    Semantic-field optimization
      & 1,500 updates (one pass over the consumed subset); learning rate
        $5\times10^{-6}$; cosine schedule; warmup ratio 0.03; KL coefficient
        0.02 against the SFT policy. \\
    Semantic-field reward
      & Direction/shape/amplitude/peak weights 0.20/0.20/0.30/0.30.
        Unparseable outputs receive $-1.5$; additional penalties enforce
        peak--amplitude, direction--amplitude, and narrative consistency. \\
    \bottomrule
  \end{tabular}
\end{table}

\begin{table}[H]
  \centering
  \caption{Forecast-utility RL (GRPO) configuration.}
  \label{tab:app-stage2-config}
  \footnotesize
  \setlength{\tabcolsep}{3.5pt}
  \begin{tabular}{p{0.26\columnwidth}p{0.66\columnwidth}}
    \toprule
    Component & Configuration \\
    \midrule
    Forecast-utility data and sampling
      & 5,500 prompts, one epoch (1,834 updates); group size 8; the same
        maximum prompt/completion lengths and rollout sampling as semantic-field RL. \\
    Forecast-utility optimization
      & Learning rate $2\times10^{-6}$; cosine schedule; warmup ratio 0.03;
        global KL coefficient 0.02 against the frozen semantic-field policy, plus a
        prompt-conditional SFT KL coefficient in $[0,0.004]$. \\
    Forecast-utility routing and gate
      & Sample skip/basic/tool with each rollout; disallow skip for holidays and
        mega-sales. Skip bypasses field parsing and uses the cached no-text
        forecast. For non-skip routes, require field reward
        $A_i\geq\max(A_i^{\mathrm{ref}}-0.08,0.10)$ and reject hard cross-field
        contradictions before invoking the frozen fusion model. \\
    Forecast utility
      & Compare candidate MAE with the cached semantic-field-policy fusion MAE and the
        no-text TSFM MAE; penalize harm beyond $1.05$ times the no-text MAE.
        Utility is clipped on a 0.25 relative-error scale. Its weight is 0.5
        in the reference run and 1.5 after the fixed step-500 diagnostic; a
        normalized route-cost penalty makes tool use more expensive than basic
        reasoning and assigns zero cost to skip. \\
    \bottomrule
  \end{tabular}
\end{table}

\end{document}